\documentclass[letterpaper]{article} 
\usepackage{aaai2026}  
\usepackage{times}  
\usepackage{helvet}  
\usepackage{courier}  
\usepackage[hyphens]{url}  
\usepackage{graphicx} 
\usepackage{natbib}  
\usepackage{caption} 
\title{GHOST: Solving the Traveling Salesman Problem on Graphs of Convex Sets}
\author{
    Jingtao Tang, Hang Ma
}
\affiliations{
    Simon Fraser University\\

    \{jingtao\_tang, hangma\}@sfu.ca
}

\usepackage{bibentry}

\usepackage[shortlabels]{enumitem}
\usepackage{amsmath,amssymb,amsfonts}
\usepackage{algpseudocode}
\usepackage[linesnumbered, ruled, vlined, noend]{algorithm2e}
\SetAlFnt{\small}
\SetAlCapFnt{\small}
\SetAlCapNameFnt{\small}
\SetArgSty{textrm}
\SetInd{0.1em}{0.5em}
\SetCommentSty{emph}
\SetAlgoSkip{}
\usepackage{etoolbox}
\makeatletter
\patchcmd{\@algocf@start}
  {-1.5em}
  {0pt}
  {}{}

\usepackage{xcolor}

\usepackage{amsthm}
\makeatletter
\def\thm@space@setup{%
 \thm@preskip=\parskip \thm@postskip=0pt
}
\makeatother
\usepackage{thmtools}
\declaretheorem{theorem}
\declaretheoremstyle[%
  spaceabove=0pt,
  spacebelow=0pt,
  headfont=\normalfont\itshape,%
  postheadspace=1em,%
  qed=\qedsymbol%
]{mystyle}
\declaretheorem[name={Proof},style=mystyle,unnumbered,
]{myproof}

\newtheorem{lemma}[theorem]{Lemma}
\newtheorem{corollary}[theorem]{Corollary}

\allowdisplaybreaks

\begin{document}

\maketitle

\begin{abstract}
We study GCS-TSP, a variant of the Traveling Salesman Problem (TSP) defined over a Graph of Convex Sets (GCS)---a powerful representation for trajectory planning that decomposes the configuration space into convex regions connected by a sparse graph. In GCS-TSP, edge costs are not fixed but depend on the specific trajectory passing through each convex region, making classical TSP methods inapplicable.
We introduce \textsc{GHOST}, a hierarchical framework that optimally solves GCS-TSP by combining combinatorial tour search with convex trajectory optimization. GHOST systematically explores tours on a complete graph induced by the GCS, using a novel abstract-path-unfolding algorithm to compute admissible lower bounds that guide best-first search at both the high level (over tours) and the low level (over feasible GCS paths realizing the tour). These bounds provide strong pruning power, reducing unnecessary optimization calls.
We prove that GHOST guarantees optimality and present a bounded-suboptimal variant for time-critical settings. Experiments show that GHOST is orders-of-magnitude faster than unified mixed-integer convex programming baseline while uniquely handling complex problems involving high-order continuity constraints and incomplete GCSs.
\begin{links}
     \link{Project page}{https://sites.google.com/view/ghost-gcs-tsp}
     \link{Code}{https://github.com/reso1/ghost}
\end{links}
\end{abstract}

%

\section{Introduction}

Robotic motion planning often relies on trajectory optimization, particularly in high-dimensional spaces subject to kinodynamic constraints. However, even a single obstacle in the configuration space can render the problem nonconvex and computationally challenging \cite{ratliff2009chomp,schulman2014motion}. A common approach is to reformulate such problems as convex programs for scalability. To this end, \citet{marcucci2023motion} proposed the \emph{Graph of Convex Sets} (GCS) framework, which decomposes the configuration space into convex regions connected via a graph structure. This representation enables transforming nonconvex trajectory planning into convex optimization subproblems while preserving feasibility and continuity.
Building on this foundation, prior work has extensively studied shortest-path problems on GCSs~\cite{marcucci2024shortest}, enabling globally optimal navigation under convex constraints. Yet, many real-world robotics tasks---such as area coverage, inspection, and structured exploration---require visiting multiple regions in a cost-effective order. This naturally leads to the \textit{Traveling Salesman Problem on a GCS} (GCS-TSP).

\begin{figure}
\centering
\includegraphics[width=\linewidth]{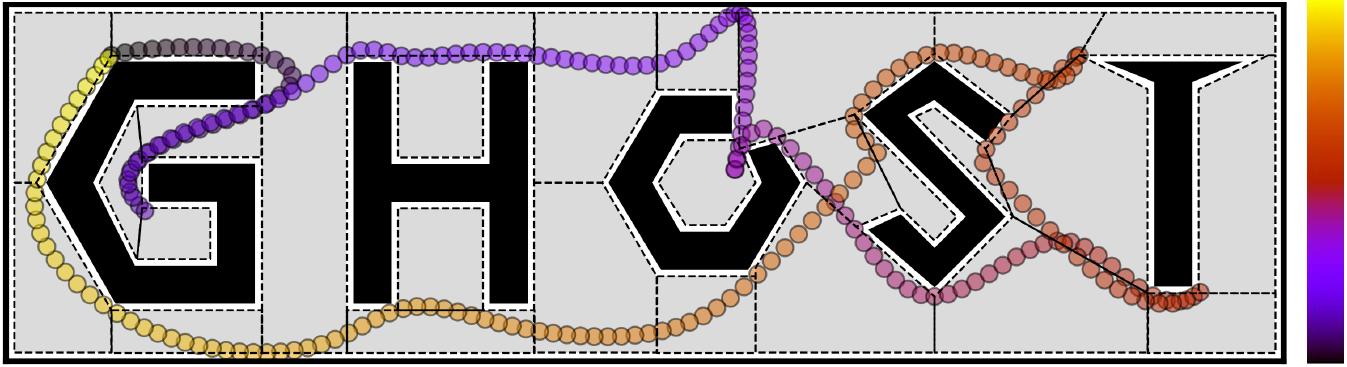}
\caption{GCS-TSP solution by GHOST. Gray polygons are convex sets. Trajectory color follows the bar (top-bottom).}
\label{fig:teaser}
\end{figure}

The input to GCS-TSP is a GCS whose vertices represent convex subsets of continuous configuration space and whose edges represent feasible transitions. The goal is to compute a tour that visits each convex set at least once and a continuous trajectory through representative points within these sets, satisfying trajectory constraints. 
This problem tightly couples combinatorial tour optimization with continuous kinodynamic feasibility. Fig.~\ref{fig:teaser} depicts an example solution generated by our method in a 2D environment. Applications include coverage and inspection planning, constrained multi-target visitation tasks, and task and motion planning.

\subsection{Related Work}

\noindent\textbf{Graphical TSP:}
The TSP seeks the shortest Hamiltonian cycle on a weighted graph, given pairwise distances between vertices. When the input graph is incomplete and may not admit a Hamiltonian cycle, the problem is typically solved on its \emph{metric closure}~\cite{cornuejols1985traveling}---a complete graph with edges weighted by shortest-path distances in the original graph. However, edge weights in a GCS are not fixed: Each transition cost depends on the specific trajectory endpoints within convex sets and may be subject to trajectory constraints. Thus, the GCS setting does not admit a static metric closure, and the TSP must be solved over dynamically evaluated edge costs.

\noindent\textbf{Combinatorial Optimization in GCSs:}
Advances in constructing collision-free convex sets~\citep{werner2024approximating,dai2024certified} have enabled efficient shortest-path solving in GCSs~\cite{marcucci2024shortest}, supporting applications such as non-Euclidean motion planning~\cite{cohn2025non}, quadrotor navigation~\cite{marcucci2024fast}, and multi-robot planning~\cite{tang2025space}. \citet{marcucci2024graphs} formulated a Mixed Integer Convex Program (MICP) for GCS-TSP, later extended to moving-target settings~\cite{philip2024mixed}, but restricted to Hamiltonian cycles on complete GCSs with poor scalability. \citet{bhat2025complete} proposed a two-level approach for 3D moving-target TSP, where the high level incrementally expands a partial GCS path, relying on problem-specific time-window assumptions, small branching factors, and piecewise-linear motion models.


\noindent\textbf{Coverage and Inspection Planning:}
TSP is central to coverage planning~\cite{galceran2013survey}, typically applied to grid graphs~\cite{gabriely2001spanning,tang2025large} or other workspace decompositions~\cite{choset1998coverage,acar2002morse}, followed by tour computation over the resulting regions. These methods often constrain robots to disk-shaped sweeps, limiting flexibility in complex environments. Recent global approaches~\cite{wu2019energy,tang2024multi} use smooth space-filling curves to cover workspaces without decomposition, but rely on predefined trajectory templates and do not consider kinodynamic constraints. Inspection planning, or the Watchman Route Problem~\cite{ntafos1992watchman}, is conceptually similar to GCS-TSP, allowing coverage by visibility in polygonal environments. However, existing methods typically 
rely on dense waypoint sampling on fixed graphs~\cite{fu2019toward} or are limited to grid-based decompositions~\cite{skyler2022solving}.

\subsection{Our Contributions}

To address the limitations of prior work, we formulate GCS-TSP and present \textsc{\textbf{G}CS-\textbf{H}ierarchical \textbf{O}ptimal \textbf{S}earch for \textbf{T}ours} (\textbf{GHOST}), a novel algorithmic framework that solves GCS-TSP optimally. GHOST employs a two-level hierarchical search: At the high level, it systematically explores TSP tours on the complete graph induced by the input GCS, ordered by lower bounds on their trajectory costs; At the low level, it searches for feasible GCS paths that realize each tour and computes the optimal trajectory for each path via convex optimization.

A central component of GHOST is a general \emph{abstract-path-unfolding} algorithm that computes admissible lower bounds on the cost of realizing a given abstract tour---or more generally, any sequence of convex sets (e.g., abstract paths) in which consecutive sets may not be directly connected in the GCS. These lower bounds can be precomputed and used to guide the best-first exploration of feasible GCS paths, significantly improving efficiency. They also serve to prioritize high-level search for pruning unpromising tour candidates early, enabling provably optimal search while avoiding unnecessary convex optimization calls.

We prove GHOST is optimal under this search strategy. We also present a bounded-suboptimal variant that trades optimality for improved scalability. 
We evaluate and demonstrate GHOST across various planning tasks in robotics.
Compared to the unified MICP baseline, GHOST is orders of magnitude faster in simple cases and uniquely capable of solving general instances previously out of reach.

\section{Problem Formulation}
In this section, we formally define GCS and GCS-TSP.

\noindent\textbf{Graph of Convex Sets (GCS):}
A GCS $G=(V,E)$ is a directed graph.
Each vertex $v\in V$ is associated with a non-empty, compact convex set $\mathcal{X}_v$.
Each edge $(u,v)\in E$ exists iff transitions are allowed from vertices $u$ to $v$. 
A \textit{path} $\pi=(v_1,\ldots,v_{|\pi|})$ is a sequence of vertices where $(v_{i}, v_{i+1})\in E$ for all $i=1,\ldots, |\pi|-1$.
A \textit{trajectory} $\tau=(\mathbf{x}_{v_1},\ldots,\mathbf{x}_{v_{|\pi|}})$ \textit{conditioned on} $\pi$ is a sequence of points where each $\mathbf{x}_{v_i}\in\mathcal{X}_{v_i}$. 
Additional convex constraints may be imposed on the trajectory, e.g., kinodynamic feasibility. 
Typically, for each edge $(v_i, v_{i+1})\in E$, constraints of the form $ (\mathbf{x}_{v_i},\mathbf{x}_{v_{i+1}})\in\mathcal{X}_{(v_i, v_{i+1})}$ may apply, where $\mathcal{X}_{(v_i, v_{i+1})}$ is a closed convex set.
Unlike in standard graphs with fixed edge weights, the cost of traversing $\pi$ depends on the optimal trajectory conditioned on it. Specifically, we define the trajectory cost as $c(\tau) = \sum_{i=1}^{|\pi|-1} c(\mathbf{x}_{v_i}, \mathbf{x}_{v_{i+1}})$, where $c(\mathbf{x}_{v_i}, \mathbf{x}_{v_{i+1}})$ is a proper, closed, convex, positive, and boundedaway-from-zero cost function.

Graph optimization problems on GCSs involve both discrete decisions (selecting sets and their visitation order) and continuous decisions (choosing trajectory points within those sets). When formulated as network flows~\cite{marcucci2024graphs}, they yield mixed-integer nonconvex programs, which are generally intractable. \citet{marcucci2023motion} introduced the ``GCS convex restriction'' trick that solves the shortest-path problem on a GCS from a given starting point $\mathbf{x}_s\in\mathcal{X}_{s}$ to a target point $\mathbf{x}_t\in\mathcal{X}_{t}$ by fixing a path $\pi$---typically simple (i.e., without vertex repetition)---and solving a convex program for the optimal trajectory conditioned on $\pi$. Paths are explored iteratively, often guided by a convex relaxation over discrete set-selection variables. Recent GCS planners~\cite{natarajan2024ixg, chia2024gcs, tang2025space} extend this idea with best-first search over feasible paths, leveraging the fact that trajectory optimization along a fixed path is convex and efficient.

\noindent\textbf{Traveling Salesman Problem on a GCS (GCS-TSP):}
GCS-TSP generalizes the classical TSP to the GCS setting. Given a GCS $G=(V,E)$, the goal is to find a \text{tour}---a closed path $\pi=(v_1,v_2,\ldots,v_{|\pi|}=v_1)$ that visits each vertex in $V$ at least once---along with its associated trajectory $\tau=(\mathbf{x}_{v_1},\mathbf{x}_{v_2},...,\mathbf{x}_{v_{|\pi|}}=\mathbf{x}_{v_1})$ through the convex sets that starts and ends at the same point, such that the trajectory cost is minimized. Formally, we define GCS-TSP as:
{\fontsize{9}{10}\selectfont\begin{align}
\min_{\pi,\mathbf{x}}\quad
& c(\tau):=\sum_{i=1}^{|\pi|-1} c(\mathbf{x}_{v_i}, \mathbf{x}_{v_{i+1}})\label{eqn:tsp:obj}\\[-6pt]
\textbf{s.t.}\quad
& \mathbf{x}_{v_1} = \mathbf{x}_{v_{|\pi|}},\\[-3pt]
&\mathbf{x}_{v_i}\in\mathcal{X}_{v_i},&\forall i=1,\ldots,|\pi|\label{eqn:tsp:vertex},\\[-3pt]
&(\mathbf{x}_{v_{i}},\mathbf{x}_{v_{i+1}})\in\mathcal{X}_{(v_{i}, v_{i+1})},&\forall i=1,\ldots,|\pi|-1\label{eqn:tsp:edge}.
\end{align}
}
In this paper, we assume the starting point and set are not fixed, following the classical TSP setting. However, our formulation and results naturally extend to cases with a fixed starting set $\mathcal{X}_s$ and point $\mathbf{x}_s$.

\section{Lower-Bound Graph and Path Unfolding}\label{sec:lbg}

In this section, we establish a static lower-bound metric for a given GCS $G=(V,E)$ to guide both levels of search in our GHOST framework. This metric is also required by the TSP subroutine (see next section), which generates tour candidates in best-first order.
The key idea is to define a lower-bound distance metric that underestimates the true trajectory cost between any vertex pair. 
Summing these lower bounds over an abstract path yields a provable lower bound cost of the trajectories conditioned on it. 
This enables GHOST to explore abstract tours in order of increasing lower-bound estimates, enabling early pruning while preserving optimality.

\noindent\textbf{Notations:} Our tour search operates over an abstract complete graph $\hat{G} = (V,\hat{E})$ induced by $G$, where $E = \{(u,v)|u,v \in V, u\neq v\}$. Each tour candidate is an \emph{abstract path} (or tour), i.e., a vertex sequence where consecutive vertices may not be adjacent in $G$. Any abstract path $\hat{\pi}=(v_1,\ldots,v_k)$ can be \textit{unfolded} into (i.e., realized by) a valid GCS path $\pi=\pi_{v_1,v_2}\circ \ldots \circ \pi_{v_{k-1},v_k}$, where each subpath $\pi_{v_i,v_{i+1}} = (v_i=w^{(i)}_1,w^{(i)}_2,\ldots, w^{(i)}_{\mathcal{K}_i}=v_{i+1})$, and $\circ$ denotes concatenation with repeated endpoints omitted.

\subsection{Lower-Bound Graph (LBG) Construction}\label{subsec:lbg_construction}

We construct a lower-bound graph (LBG) to efficiently estimate path costs.
Unlike standard graphs where edge costs provide meaningful path estimates, the cost of traversing a GCS edge $(u,v)\in E$ depends on trajectory endpoints $\mathbf{x}_u \in \mathcal{X}_u$ and $\mathbf{x}_v \in \mathcal{X}_v$, which are unconstrained unless the full path is fixed. As a result, edge costs in $G$ yield loose lower bounds. Instead, we define the LBG on each triplet $(u,v,w)$, representing a vertex $v\in V$ along with its predecessor $u\in V$ and successor $w\in V$, allowing for a tighter lower bound that captures the “volume” of set $\mathcal{X}_v$.

\begin{figure}[t]
\centering
\includegraphics[width=0.9\columnwidth]{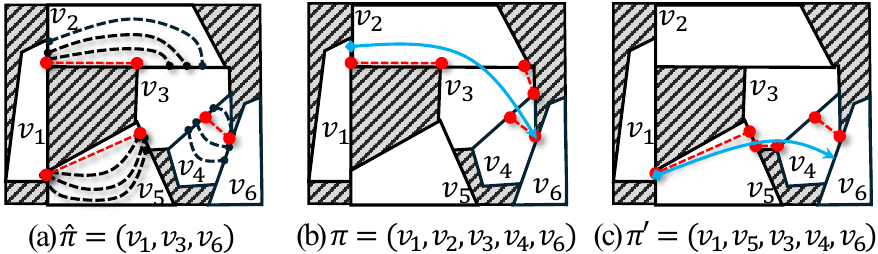}
\caption{LBG and path unfolding. (a) The optimal trajectory (red dashed line) yielding lower-bound cost $lb_p$ for any trajectories (black dashed curves) conditioned on each possible $p$ along $\hat\pi$. (b)(c) Trajectories $\tau$ and $\tau'$ (blue solid curves) conditioned on $\pi$ and $\pi'$ unfolded from $\hat\pi$, respectively.}
\label{fig:lbg}
\end{figure}

Given a GCS $G=(V,E)$, the LBG $\mathcal{H} = (\mathcal{P},\mathcal{F})$ is a directed hypergraph. A triplet $(u,v,w)\in\mathcal{P}$ is a passage formed from every pair of adjacent edges $(u,v),(v,w)\in E$, and a hyperedge $(p,p')\in \mathcal{F}$ connects each $p=(\cdot,u,v)$ to $p'=(u,v,\cdot)$. Each triplet $p=(u,v,w)$ is labeled with a lower-bound cost $lb_p$, computed as the optimal trajectory cost conditioned on $p$ via a convex program. Since this program relaxes any constraints on $\mathbf{x}_u\in\mathcal{X}_u$ and $\mathbf{x}_w\in\mathcal{X}_w$, $lb_p$ never overestimates the cost portion $c(\mathbf{x}_u,\mathbf{x}_v)+c(\mathbf{x}_v,\mathbf{x}_w)$ for any trajectory conditioned on a path using triplet $p$. 
Fig.~\ref{fig:lbg}-(a) demonstrates an example of $lb_p$.
The LBG has also been used in~\citet{natarajan2024ixg} to obtain an admissible heuristic for accelerating A$^*$-like search in GCSs.

\subsubsection{Properties of LBG:} 
Any path $\pi$ on $G$ can be equivalently written as $(p_1,\ldots,p_k)$, where each $p_i=(u_i,v_i,w_i)$, and $\pi=(u_1,v_1 =u_2,w_1=v_2=u_3,\ldots,w_{k-2}=v_{k-1}=u_k,w_{k-1}=v_k,w_k)$. We define $\mathcal{L}(\pi)$ as the cost accumulating the lower-bound costs of triplets along $\pi$:
{\fontsize{9}{10}\selectfont\begin{align*}
\mathcal{L}(\pi)=\sum_{p\text{ along }\pi}lb_p+\begin{cases}
0, \text{ if $\pi$ is open (i.e., $u_1 = w_k$)}\\
lb_{(v_k,u_1,v_1)}, \text{ otherwise ($\pi$ is closed)}
\end{cases}
\end{align*}}
Theorem 3 in \citet{natarajan2024ixg} show that for any trajectory $\tau$ from $\mathcal{X}_u$ to $\mathcal{X}_{w'}$ conditioned on an open path $\pi$, we have $\mathcal{L}(\pi^*_{u,w'}) \leq \mathcal{L}(\pi) \leq c(\tau)$, where $\pi^*_{u,w'}$ is the \textit{lower-bound-cost-minimal path} from $\mathcal{X}_u$ to $\mathcal{X}_{w'}$.
Theorem~\ref{theo:lb} generalizes this result to any path unfolded from an abstract path.
\begin{theorem}[Lower-Bound Path Cost]\label{theo:lb}
For any abstract path $\hat{\pi}$ on $\hat{G}$ and the lower-bound-cost-minimal $\pi^*$ unfolded from it, $\mathcal{L}(\pi^*)\leq \mathcal{L}(\pi)\leq c(\tau)$ holds for any trajectory $\tau$ conditioned on an arbitrary $\pi$ unfolded from $\hat{\pi}$.
\end{theorem}
\begin{myproof}
The first inequality follows by definition. The second follows from Theorem 3 in \citet{natarajan2024ixg} for open paths and extends to closed paths by noting that $\mathcal{L}(\pi)$ is the cost of a relaxed trajectory optimization conditioned on $\pi=(p_1,\ldots,p_k)$ by ignoring ``connection constraints'' between consecutive triplets $p_i\rightarrow p_{i+1}$. The same argument applies to a closed path $\pi$, where the additional triplet cost term similarly ignores these constraints for the wraparound connections $p_{k-1}\rightarrow p_k=p_1$ and $p_k=p_1\rightarrow p_2$. 
\end{myproof}

\subsection{Path Unfolding}
We present our (abstract-)path-unfolding algorithm, which uses a multi-label A$^*$-like best-first search~\cite{grenouilleau2019multi} to enumerate all paths unfolded from a given abstract path $\hat{\pi}=(v_1,v_2,\ldots,v_k)$ on $\hat{G}$ in non-decreasing order of their lower-bound costs. Each search state consists of a triplet $p$ and a label $\ell$, indicating progress toward the next endpoint $v_{\ell+1}$. 
To allow for cyclic and next-best paths to a state, each search node $n$ consists of its path $n.\pi$ (instead of a triplet) and label $n.\ell$. The node also stores the current cost $n.g$ and an estimated final cost $n.f = n.g + h_{\hat\pi}(p, n.\ell)$, where $h_{\hat\pi}(p, n.\ell)$ is an admissible heuristic underestimating the cost to sequentially visit all remaining endpoints $v_{\ell+1}, \ldots, v_k$.
Fig.~\ref{fig:lbg}-(bc) shows an example in which the optimal trajectory conditioned on $\pi$ can have a smaller cost than $\pi'$, even when $\mathcal{L}(\pi) < \mathcal{L}(\pi')$.
This motivates exhaustively exploring all unfolded paths for $\hat{\pi}$ to guarantee finding the one that yields the optimal trajectory.

\begin{figure}
\centering
\includegraphics[width=0.73\columnwidth]{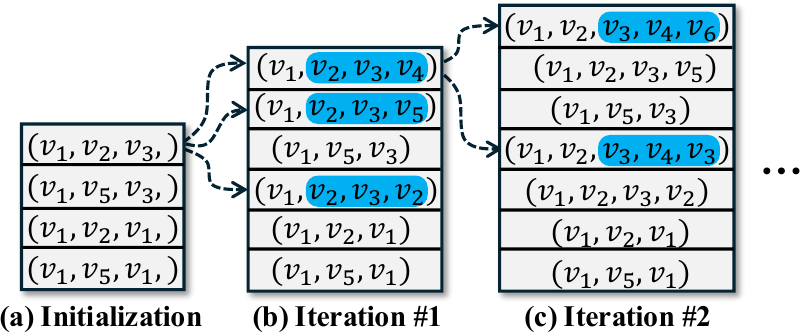}
\caption{Illustration of Alg.~\ref{alg:bfs} for $\hat\pi=(v_1,\ldots)$ on the GCS from Fig.~\ref{fig:lbg}: OPEN evolution (nodes in gray blocks sorted top-bottom in non-decreasing $f$), node expansions (dashed arrows), and appended triplets (blue).}
\label{fig:path_unfolding}
\end{figure}

\noindent\textbf{Pseudocode (Alg.~\ref{alg:bfs})}
The search maintains an OPEN list of nodes $n$, prioritized by $n.f$ (Line~\ref{alg:bfs:open_list}). Nodes are initialized for every triplet $p$ that matches the prefix of $\hat{\pi}$, with their labels set according to how many initial vertices of $\hat{\pi}$ are matched (Lines~\ref{alg:bfs:prefix}-\ref{alg:bfs:all_roots}). At each step, $n$ with the lowest $n.f$ is popped (Line~\ref{alg:bfs:pop}). For an open $\hat{\pi}$, if $n.\pi$ ends at $v_k$ and $n.\ell = k$, the search outputs $n.\pi$ as the next-best unfolded path (Line~\ref{alg:bfs:open_leaf}). For a closed $\hat{\pi}$, if $n.\pi$ starts and ends at the same triplet $(\cdot, v_k, \cdot)$ and $n.\ell = k$, the search outputs $n.\pi$ as the next-best unfolded path (Line~\ref{alg:bfs:closed_leaf}), with $n.g$ correctly capturing the wraparound triplet cost. The search then branches on all triplets $p$ consistent with the suffix of $n.\pi$. Fig.~\ref{fig:path_unfolding} shows an example. Each child node $n_c$ increments the label if appending $p$ reaches endpoint $v_{n.\ell+1}$ (Line~\ref{alg:bfs:update_label}), updates $n_c.g$ by adding $lb_p$, and sets $n_c.f$ as the sum of $g$ and an admissible heuristic value $h_{\hat\pi}$ (Line~\ref{alg:bfs:gf_cost}). Optionally, nodes with $n_c.f > \bar{c}$ are pruned (Line~\ref{alg:bfs:prune}), supporting efficient hierarchical search in GHOST.
We employ a multi-label heuristic inspired by~\citet{li2021lifelong,zhong2022optimal} as defined in Lemma~\ref{lemma:ml_heur}.
Fig.~\ref{fig:path_unfolding} illustrates Alg.~\ref{alg:bfs} with a concrete example.
\begin{lemma}\label{lemma:ml_heur}
The multi-label heuristic value $h_{\hat{\pi}}(p,n.\ell)=\mathcal{L}(\pi^*_{w,v_{\ell+1}})+\sum_{i=\ell+1}^{k-1}\mathcal{L}(\pi^*_{v_{i},v_{i+1}})$ never overestimates the actual lower-bound cost-to-go for any path $\pi$ unfolded from $\hat{\pi}=(v_1,v_2,\ldots,v_k)$ via $p=(u,v,w)$ with label $n.\ell$.
\end{lemma}
\begin{myproof}
Consider any open path $\pi=\pi_{v_1,v_2}\circ \ldots \circ \pi_{v_{k-1},v_k}$ unfolded from $\hat{\pi}$, where each subpath $\pi_{v_i,v_{i+1}} = (v_i=w^{(i)}_1,w^{(i)}_2,\ldots, w^{(i)}_{\mathcal{K}_i}=v_{i+1})$.
Now, assume the triplet $p=(u,v,w)$ at label $n.\ell$ is followed by some $p'=(v,w,\cdot)$.
The cost-to-go from $p$ can be decomposed as $lb_{p'}+\mathcal{L}(\pi_{w,v_{\ell+1}})+\sum_{i=\ell+1}^{k-1}\left(lb_{(w^{(i)}_{\mathcal{K}_i-1},v_i,w^{(i+1)}_{2})} + \mathcal{L}(\pi_{v_{i},v_{i+1}})\right)$, which is at least $\mathcal{L}(\pi^*_{w,v_{\ell+1}})+\sum_{i=\ell+1}^{k-1}\mathcal{L}(\pi^*_{v_{i},v_{i+1}}) = h_{\hat{\pi}}(p,n.\ell)$. For closed paths, the inequality still holds with an additional non-negative term $lb_{(w^{(k-1)}_{\mathcal{K}_{k-1}-1},v_k,w^{(1)}_2)}$ on the left. In summary, $h_{\hat{\pi}}$ aggregates the minimum possible subpath costs but does not account for the triplet costs for $p'$ and trajectory endpoints in the actual cost-to-go.
\end{myproof}
\begin{algorithm}[t]
\footnotesize
\DontPrintSemicolon
\linespread{0.3}\selectfont
\caption{(Next-)Best-First Path Unfolding}\label{alg:bfs}
\SetKwInput{KwInput}{Input}
\KwInput{abstract path $\hat{\pi}=(v_1,\ldots,v_k)$, pruning cost $\bar{c}$}
$\text{OPEN} \gets \emptyset$\Comment{prioritizing nodes by their $n.f$}\label{alg:bfs:open_list}\;
\ForEach{$p=(v_1,u,w)\in \mathcal{P}$\label{alg:bfs:prefix}}{
    create node $n$ with $n.\ell\gets 1$, $n.\pi\gets p$, and $n.g\gets lb_p$\;
    \If{$u=v_2$}{
        $n.\ell\gets 2$\;
        \If{$w=v_3$}{
            $n.\ell\gets 3$\;
        }
    }
    $n.f\gets n.g+h_{\hat{\pi}}(p,n.\ell)$ and $\text{OPEN} \gets \text{OPEN}\cup\{n\}$\label{alg:bfs:all_roots}\;
}
\While{$\text{OPEN}\neq\emptyset$}{\label{alg:bfs:main_loop}
    $n\gets$ OPEN$.pop()$\label{alg:bfs:pop}\;
    \If{$v_1\neq v_k$ \textbf{and} $n.\ell =k$ \textbf{and} $n.\pi$ ends at $(\cdot,\cdot,v_k)$}{
        \textbf{yield} $n.\pi$ as the next-best unfolded (open) path\label{alg:bfs:open_leaf}\;
    }
    \If{$v_1=v_k$ \textbf{and} $n.\ell=k$ \textbf{and} $n.\pi$ starts and ends at the same $(\cdot,v_k,\cdot)$}{
        \textbf{yield} $n.\pi$ as the next-best unfolded (closed) path\label{alg:bfs:closed_leaf}\;
    }
        \ForEach{$p=(u,v,w)\in \mathcal{P}$ where $n.\pi$ ends at $(\cdot,u,v)$\label{alg:bfs:branch}}{
            $n_c\gets$ new child node of $n$ with $n_c.\pi$ appending $p$ to $n.\pi$\;
            \eIf{$w=v_{n.\ell+1}$}            
            {$n_c.\ell\gets n.\ell+1$\label{alg:bfs:update_label}\;}
            {$n_c.\ell \gets n.\ell$\;}
            $n_c.g\gets n.g + lb_p$, and $n_c.f\gets n_c.g + h_{\hat{\pi}}(p,n.\ell)$\label{alg:bfs:gf_cost}\;
            \If{$n_c.f\leq \bar{c}$}{\label{alg:bfs:prune}
                $\text{OPEN} \gets \text{OPEN}\cup\{n_c\}$\;\label{alg:bfs:update_OPEN}
            }
    }
}
 \end{algorithm}

\section{Restricted-TSP over Abstract Triplets}\label{subsec:ilp_rtsp}
In this section, we describe the key subroutine used to compute the next best abstract tour on the induced complete graph $\hat{G}$. The procedure solves the \textit{Restricted-TSP} (RTSP)~\cite{hamacher1985k} on $\hat{G}=(V,\hat{E})$. A feasible RTSP solution is a simple abstract tour $\hat{\pi}$ that visit all vertices.
Unlike standard TSP, RTSP specifies two disjoint abstract edge sets: $\hat{E}^+\subseteq\hat{E}$ (edges that must be included) and $\hat{E}^-\subseteq\hat{E}$ (edges that must be excluded). Standard TSP is a special case with $\hat{E}^+=\hat{E}^-=\emptyset$.

While the ILP formulation for TSP~\cite{miller1960integer} can be adapted to RTSP with a static pairwise distance metric over vertices, this is insufficient for GCS, where edge costs do not provide meaningful estimates or capture the ``volume'' of tour endpoint sets. Instead, we leverage the triplet-based distance metric established above and reformulate RTSP over all abstract triplets $\hat{p}\in\hat{\mathcal{P}}$. Each $\hat{p}=(u,v,w)$ is formed from every pair of abstract edges $(u,v),(v,w)\in\hat{E}$, with a lower-bound cost defined as $b(\hat{p})=\frac{1}{2}\mathcal{L}(\pi^*_{u,v}) + b_\text{mid}(\hat{p}) + \frac{1}{2}\mathcal{L}(\pi^*_{v,w})$, where
{\fontsize{9}{10}\selectfont\begin{align*}
b_\text{mid}(\hat{p})=\begin{cases}
lb_{(u,v,w)},&\text{if } (u,v),(v,w) \in E \\ 
\min\limits_{p=(u,v,\cdot)\in\mathcal{P}}lb_p,&\text{if } (u,v)\in E, (v,w)\notin E\\
\min\limits_{p=(v,w,\cdot)\in\mathcal{P}}lb_p,&\text{if } (u,v)\notin E, (v,w)\in E\\
\min\limits_{p=(\cdot,v,\cdot)\in\mathcal{P}}lb_p,&\text{if } (u,v),(v,w)\notin E
\end{cases}
\end{align*}}
The lower-bound path cost $\mathcal{L}(\pi^*_{u,v})$ is typically zero when $(u,v)\in E$ (and thus $\mathcal{X}_u,\mathcal{X}_v$ are adjacent sets), unless with complex vertex or edge constraints.
For notational simplicity, we use the same symbol $\mathcal{L}$ to denote the cumulative cost of the abstract triplets along an abstract path $\hat\pi$, that is, we define $\mathcal{L}(\hat\pi)=\sum_{\hat{p} \text{ along } \hat{\pi}}b(\hat{p})$ for any abstract path $\hat\pi$.
Theorem~\ref{theo:tour_lb} establishes $\mathcal{L}(\hat\pi)$ as a valid lower-bound tour cost.
\begin{theorem}[Lower-Bound Tour Cost]\label{theo:tour_lb}
For an abstract tour $\hat{\pi}$ on $\hat G$, $\mathcal{L}(\hat\pi) \leq \mathcal{L}(\pi) \leq c(\tau)$ holds for any closed path $\pi$ unfolded from $\hat{\pi}$ and any trajectory $\tau$ conditioned on $\pi$.
\end{theorem}
\begin{myproof}
The second inequality follows Theorem~\ref{theo:lb}. 
Consider $\hat{\pi}=(v_1,\ldots,v_{k})$ and an unfolded $\pi=\pi_{v_1,v_2}\circ\ldots\circ \pi_{v_{k-1},v_k}$, where each subpath $\pi_{v_i,v_{i+1}} = (v_i=w^{(i)}_1,w^{(i)}_2,\ldots, w^{(i)}_{\mathcal{K}_i}=v_{i+1})$. We decompose
$\mathcal{L}(\pi)=\sum_{i=1}^k \mathcal{L}(\pi_{v_i,v_{i+1}})+ \sum_{i=1}^k lb_{(w^{(i-1)}_{\mathcal{K}_{i-1}-1},v_i,w^{(i)}_2)}$,
where $v_{k+1} = v_1$ (wraparound) and $w^{(0)}_{\mathcal{K}_0} = w^{(k-1)}_{\mathcal{K}_{k-1}-1}$ for notation convenience.
As each $\mathcal{L}(\pi_{v_i,v_{i+1}})\geq \mathcal{L}(\pi^*_{v_i,v_{i+1}})$ by definition and $lb_{(w^{(i-1)}_{\mathcal{K}_{i-1}-1},v_i,w^{(i)}_2)}\geq b_\text{mid}(v_{i-1},v_i,v_{i+1})$ by construction, summing over all $i=1,...,k$ concludes the proof.
\end{myproof}
Unlike classical TSP ILP, which uses variables for edges, our RTSP ILP introduces binary variables $y_{\hat{p}}$ for triplets $\hat{p}\in\hat{\mathcal{P}}$, minimizing the lower-bound tour cost:
{\fontsize{9}{10}\selectfont\begin{align}
\min_{y_{\hat{p}}}\quad
&\sum_{\hat{p}\in \hat{\mathcal{P}}}b(\hat{p})y_{\hat{p}}\label{eqn:triplet_rtsp:obj}\\[-6pt]
\textbf{s.t.}\quad
&\sum_{\hat{p}=(\cdot,v,\cdot)\in\hat{\mathcal{P}}}y_{\hat{p}}= 1,& \forall v\in V\label{eqn:triplet_rtsp:cover}\\[-3pt]
&\sum_{z\neq v}y_{zuv}=\sum_{w\neq u}y_{uvw},& \forall (u,v)\in\hat{E}\label{eqn:triplet_rtsp:flow_csv}\\[-3pt]
&\sum_{\hat{p}\in\hat{\mathcal{P}}^\text{ind}_{e}}y_{\hat{p}}=2,& \forall e=(u,v)\in\hat{E}^+\label{eqn:triplet_rtsp:inc}\\[-3pt]
&y_{\hat{p}}=0,& \forall \hat{p}\in\hat{\mathcal{P}}^\text{ind}_{e},e\in\hat{E}^-\label{eqn:triplet_rtsp:exc}
\end{align}}
where $\hat{\mathcal{P}}^\text{ind}_{e=(u,v)}=\{\hat{p}\in\hat{\mathcal{P}}\,|\,\hat{p}=(\cdot,u,v)\text{ or }\hat{p}=(u,v,\cdot)\}$ denotes all triplets induced by $(u,v)$.
Eqn.~(\ref{eqn:triplet_rtsp:cover}) ensures that each $v\in V$ is visited exactly once.
Eqn.~(\ref{eqn:triplet_rtsp:flow_csv}) enforces flow conservation.
Eqn.~(\ref{eqn:triplet_rtsp:inc}-\ref{eqn:triplet_rtsp:exc}) enforce inclusion and exclusion constraints.
Whenever a new subtour $C$ is found in the solution, we add a subtour elimination constraint similar to the DFJ constraint~\cite{dantzig1954solution}:
{\fontsize{9}{10}\selectfont\begin{align}
\sum_{\hat p\in C}y_{\hat{p}}\leq |C|-1
\end{align}}
and re-solve until only a single tour remains.

\section{The GHOST Framework}
In this section, we present the GHOST framework for solving GCS-TSP. GHOST leverages the LBG $\mathcal{H}$ and path unfolding to compute lower-bound costs that guide and prune the search at both levels. At the high level, GHOST explores abstract tours $\hat{\pi}$ on the induced complete graph $\hat{G}$ in non-decreasing order of their lower-bound costs. At the low level, for each abstract tour, it explores unfolded paths $\pi$ from $\hat{\pi}$, also in non-decreasing order of lower-bound cost, and computes the optimal trajectory for each $\pi$ via convex optimization. The resulting trajectory cost is used to prune any abstract tours or paths whose lower-bound costs are no smaller, guaranteeing optimality as formally shown later.

\subsection{High-Level Tour Search}
\begin{figure}[t]
\centering
\includegraphics[width=\linewidth]{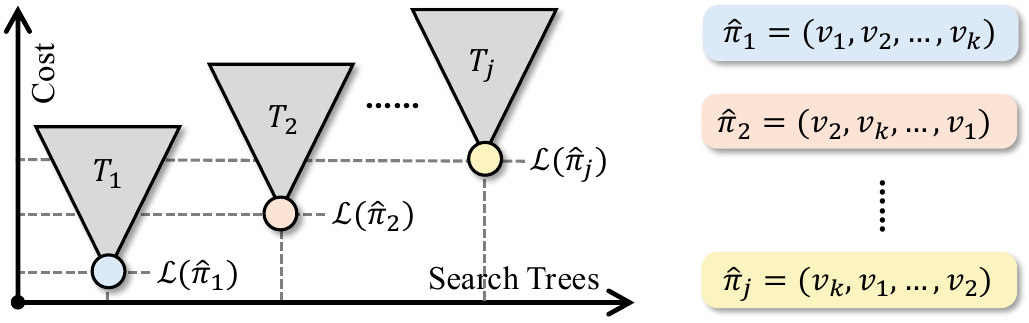}
\caption{High level of GHOST. Each $T_i$ denotes a low-level search tree that explores unfolded paths for abstract tour $\hat\pi_i$. The search forest is pruned at $T_j$ where $\mathcal{L}(\hat\pi_j)$ exceeds the current best trajectory cost.}
\label{fig:ghost}
\end{figure}

At the high level, GHOST generalizes the \textit{Lawler-Murty} procedure~\cite{murty1968algorithm,lawler1972procedure}, originally developed to find the $K$-best solutions to TSP and related combinatorial optimization problems \cite{hamacher1985k,ren2023cbss}. The procedure can be conceptualized as a tree search where each node corresponds to an RTSP instance.
Fig.~\ref{fig:ghost} illustrates the high-level search.

\noindent\textbf{Pseudocode} (Alg.~\ref{alg:search}): Given a GCS $G$, each search node $n$ corresponds to solving an RTSP instance on the induced complete graph $\hat{G}$ with specified inclusion $\hat{E}^+$ and exclusion $\hat{E}^-$ edge sets, and stores the resulting optimal abstract tour $n.\hat\pi$.
Function $\texttt{EvalNode}(n)$ unfolds $n.\hat\pi$ and evaluates the trajectory cost. GHOST performs a best-first search with an OPEN list prioritized by the lower-bound tour cost $n.\underline{c}$ (Line~\ref{alg:search:open_list}).
The root node $r$ is initialized with the unrestricted TSP, $n^*$ tracks the best node found so far (Lines~\ref{alg:search:root}-\ref{alg:search:dummy_best}). In each iteration, the node $n$ with the smallest $n.\underline{c}$ is popped (Line~\ref{alg:search:pop}).
If $n.\underline{c}$ is no smaller than the cost of the current best trajectory $c(n^*.\tau)$ (Line~\ref{alg:search:llb_ret}), GHOST returns $n^*.\tau$ as optimal (Line~\ref{alg:search:return}).
Otherwise, GHOST calls $\texttt{EvalNode}$ to unfold $n.\hat\pi$ and compute $c(n.\tau)$ (Line~\ref{alg:search:evaluate_node}).
If $c(n.\tau)$ improves upon $c(n^*.\tau)$, $n^*$ is updated (Line~\ref{alg:search:update_n_opt}).
For each edge $e_i$ in $n.\hat{\pi}$, GHOST creates a child node $n_c$, (Lines~\ref{alg:search:exp_child_loop}-\ref{alg:search:create_child}) by adding new constraints to its parent: It excludes $e_i$ (Line~\ref{alg:search:update_Eout}) and includes all edges with indices less than $i$ (Line~\ref{alg:search:update_Ein}). If the resulting  RTSP instance is feasible, $n_c$ is inserted into OPEN with its tour and lower bound (Lines~\ref{alg:search:no_feasible_rtsp}-\ref{alg:search:add_child}).

\begin{algorithm}[t]
\footnotesize
\DontPrintSemicolon
\linespread{0.3}\selectfont
\caption{GHOST for GCS-TSP}\label{alg:search}
\SetKwInput{KwInput}{Input}
\KwInput{GCS $G=(V,E)$}
$r\gets$ a root node with $r.\hat{E}^+ = r.\hat{E}^- = \emptyset$ and empty $r.\tau$\label{alg:search:root}\;
$r.\hat\pi\gets$ optimal TSP tour and $r.\underline{c} \gets \mathcal{L}(r.\hat{\pi})$\label{alg:search:root_rtsp}\;
initialize the best $n^*$ as a dummy node with $n^*.\tau=\infty$\label{alg:search:dummy_best}\;
OPEN $\gets \{r\}$\Comment{prioritizing nodes by their $n.\underline{c}$}\label{alg:search:open_list}\;
\While{{\normalfont OPEN} $\neq\emptyset$}{
    $n\gets$ OPEN$.pop()$\label{alg:search:pop}\;
    \If{$n.\underline{c} \geq c(n^*.\tau)$}{\label{alg:search:llb_ret}
        \textbf{break}\;
    }
    \texttt{EvalNode}($n,c(n^*.\tau)$)\label{alg:search:evaluate_node}\;
    update $n^*$ to $n$ if $c(n.\tau)<c(n^*.\tau)$\label{alg:search:update_n_opt}\;
    \ForEach{edge $e_i$ in $n.\hat\pi = (1, \ldots, e_{|n.\hat\pi|})$}
    {\label{alg:search:exp_child_loop}
        $n_c\gets$ new child node of $n$\label{alg:search:create_child}\;
        $n_c.\hat{E}^-\gets \{e_i\}\cup n.\hat{E}^-$\label{alg:search:update_Eout}\;
        $n_c.\hat{E}^+\gets \{e_j\}_{j=1}^{i-1}\cup n.\hat{E}^+$\label{alg:search:update_Ein}\;

        \If{{\normalfont a feasible RTSP tour exists for $(n_c.\hat{E}^+,n_c.\hat{E}^-)$\label{alg:search:no_feasible_rtsp}}}{
            $n_c.\hat\pi\gets$ optimal RTSP tour for $(n_c.\hat{E}^+,n_c.\hat{E}^-)$\label{alg:update_node:rtsp_solve}\;
        $n_c.\underline{c}\gets \mathcal{L}(n_c.\hat{\pi})$ and  OPEN $\gets$ OPEN $\cup\{n_c\}$\label{alg:search:add_child}\;
        }
        
    }
}
\Return optimal GCS-TSP trajectory $n^*.\tau$\;\label{alg:search:return}
\SetKwFunction{FMain}{EvalNode}
\SetKwProg{Fn}{Function}{:}{}
\Fn{\FMain{$n,\bar{c}$}}{\label{alg:update_node:begin}
    $c(n.\tau)\gets \infty$\;
    \ForEach{$\pi\gets$ next-best path unfolded from $\hat\pi$ (by calling Alg.~\ref{alg:bfs} with pruning cost $\bar{c}$)}{\label{alg:update_node:next_best_GTSP_tour}
        $\tau\gets$ optimal trajectory conditioned on $\pi$\label{alg:update_node:opt_traj}\;       \If{$\mathcal{L}(\pi)\geq c(n.\tau)$\label{alg:update_node:no_better_path}}{\textbf{break}\label{alg:update_node:break}}
        update $n.\tau$ to $\tau$ if $c(\tau)<c(n.\tau)$\label{alg:update_node:update_ntau}\;
    }
}
\end{algorithm}

\subsection{Tour Unfolding and Trajectory Optimization}
The low-level \texttt{EvalNode} function calls our path-unfolding algorithm (Alg.~\ref{alg:bfs}) to generate path $\pi$ unfolded from $\hat\pi$, considering each in non-decreasing order of $\mathcal{L}(\pi)$ (Line~\ref{alg:update_node:next_best_GTSP_tour}) and pruning once $\mathcal{L}(\pi)$ exceeds the best trajectory cost $c(n.\tau)$ among all previously considered paths for $\hat{\pi}$ (Lines~\ref{alg:update_node:no_better_path}-\ref{alg:update_node:break}). For each $\pi$, the function computes the optimal trajectory $\tau$ conditioned on $\pi$ via convex optimization (Line~\ref{alg:update_node:opt_traj}) and updates $n.\tau$ if a lower cost is found (Line~\ref{alg:update_node:update_ntau}).

The convex optimization follows “GCS convex restriction” of~\citet{marcucci2023motion}, but with a key extension: Our formulation allows $\pi$ to revisit vertices and edges multiple times (which can occur when $G$ is incomplete). We handle this by introducing separate variables for each occurrence of a vertex and defining edge traversal constraints according to the specific sequence in $\pi$. To improve efficiency, we cache the optimal trajectory for each $\pi$ since the same $\pi$ may be unfolded from different abstract tours.

\subsection{Bounded-Suboptimal $\epsilon$-GHOST}
We introduce a bounded-suboptimal variant, $\epsilon$-GHOST, which trades optimality for computational efficiency. Given a parameter $\epsilon \in [0,1)$, we modify GHOST in two ways: On Line~\ref{alg:search:llb_ret}, the termination condition is relaxed from $c(n^*.\tau)-n.\underline{c}<0$ to $c(n^*.\tau)-n.\underline{c}\leq \epsilon \cdot c(n^*.\tau)$. In \texttt{EvalNode}, Alg.\ref{alg:bfs} is called with a pruning cost $(1-\epsilon)\cdot c(n^*.\tau)$ instead of $c(n^*.\tau)$ (Line~\ref{alg:update_node:next_best_GTSP_tour}), and the pruning condition is updated to $\mathcal{L}(\pi) \geq (1-\epsilon)\cdot c(n.\tau)$ instead of $\mathcal{L}(\pi) \geq c(n^*.\tau)$ (Line~\ref{alg:update_node:no_better_path}).
Setting $\epsilon= 0$ recovers the original, optimal GHOST.

\subsection{Properties of ($\epsilon$-)GHOST}

As in \citet{van1999solving,ren2023cbss}, the Lawler-Murty procedure at the high level of GHOST guarantees to explore TSP tours in non-decreasing order of their lower-bound costs. Theorem~\ref{theo:bounded_opt} and Corollary~\ref{theo:opt} establish the (bounded-sub)optimality of ($\epsilon$-)GHOST. 

\begin{theorem}\label{theo:bounded_opt}
$\epsilon$-GHOST produces a GCS-TSP solution with cost at most $\frac{1}{1-\epsilon}$ times the optimal cost.
\end{theorem}
\begin{myproof}
Consider any node $n_i$ popped at iteration $i$, where $\texttt{EvalNode}$ is called with pruning cost $(1-\epsilon) \cdot c(n_i^*.\tau)$, and $n_i^*$ stores the best node over all previous iterations. Let $\bar{\tau}_i$ denote any pruned trajectory conditioned on some path $\pi$ unfolded from $n_i.\hat{\pi}$. Such a trajectory is pruned either (1) in Alg.~\ref{alg:bfs} if $\mathcal{L}(\pi)\geq (1-\epsilon) \cdot c(n_i^*.\tau)$, or (2) in $\texttt{EvalNode}$ if $\mathcal{L}(\pi) \geq (1-\epsilon) \cdot c(n_i^*.\tau)$. Since unfolded paths are considered in non-decreasing order of $\mathcal{L}(\pi)$, in either case we have $c(\bar{\tau}_i)\overset{\text{Theorem~\ref{theo:lb}}}\geq \mathcal{L}(\pi) \geq (1-\epsilon) \cdot c(n^*.\tau)$, where $n^*.\tau$ is the best trajectory across all iterations. 

When $\epsilon$-GHOST terminates and returns $n^*.\tau$, the termination condition ensures $n.\underline{c} \geq (1-\epsilon) \cdot c(n^*.\tau)$. Since the \textit{Lawler-Murty} procedure explores tours in non-decreasing order of their lower-bound costs, any pruned trajectory $\hat\tau$ conditioned on some path unfolded from an unexplored tour $\hat{\pi}$ satisfies $c(\bar{\tau})\overset{\text{Theorem~\ref{theo:tour_lb}}}\geq \mathcal{L}(\hat{\pi}) \geq n.\underline{c}\geq(1-\epsilon)\cdot c(n^*.\tau)$.

Therefore, $c(n^*.\tau) \leq \frac{1}{1-\epsilon}\cdot c(\bar{\tau})$ for any pruned trajectory $\bar{\tau}$, which is at most $\frac{1}{1-\epsilon}$ times the optimal cost.
\end{myproof}

\begin{corollary}\label{theo:opt}
GHOST returns an optimal GCS-TSP solution.
\end{corollary}
\begin{myproof}
This follows from Theorem~\ref{theo:bounded_opt} with $\epsilon=0.0$.
\end{myproof}

\subsection{Implementation Details and GHOST Variants}

GHOST inherently considers all paths, including cyclic ones, as required by rare, complex constraints (e.g.,~\citet{natarajan2024ixg}). 
Since such cases are uncommon, many GCS shortest-path planners treat cyclic paths as optional and typically restrict to simple paths.
Our implementation similarly restricts path unfolding to simple subpaths between tour endpoints, which limits (bounded-sub)optimality to trajectories in this class (each vertex visited at most $|V|$ times), but yields improved efficiency.

GHOST sequentially explores tours, fully evaluating all the unfolded paths for one tour before considering the next. While it is possible to compare paths from different tours in parallel, we found it offers little practical benefit, as pruning often eliminates most paths from each tour---especially when $\epsilon$ is used.

Finally, extending the current triplet-based LBG and the RTSP ILP to higher-order tuples (e.g., 4-tuples) could yield tighter bounds. However, we observed that this incurs substantial computational overhead with little improvement.

\begin{figure*}
\centering
\includegraphics[width=\linewidth]{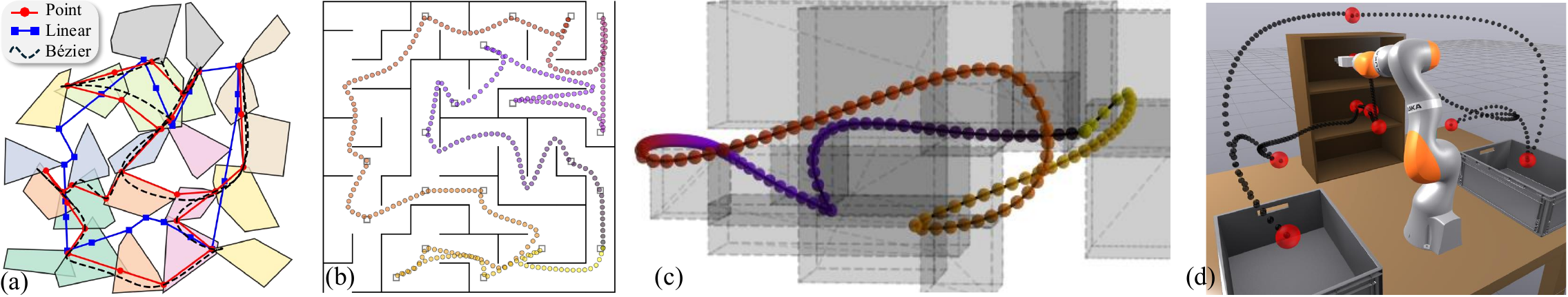}
\caption{A collection of GCS-TSP problem solutions produced by GHOST. (a) GCS-TSP with 23 convex sets (colored polygons) and 50 edges on Point-GCS, Linear-GCS, and Bézier-GCS. (b) Ground vehicle inspection planning for target location sets (squares) in a 2D maze. (c) Quadrotor coverage planning in a 3D obstacle-rich (gray cuboids) bounded region. (d) Task and motion planning for a 7-DoF manipulator, where black and red spheres indicate the trajectory and task poses, respectively.}
\label{fig:demo}
\end{figure*}

\section{Empirical Evaluation}
This section presents our experimental results on an \textit{Apple}\textsuperscript{\textregistered} M4 CPU machine with 16GB RAM.
We use \textit{Drake}~\cite{drake} with \textit{Gurobi} solver for solving the GCS convex restriction and the RTSP ILP.

\subsection{Experiment Setup}
To evaluate our \textbf{GHOST} framework with LBG (Alg.~\ref{alg:search}) for GCS-TSP, we compare against the following baselines:

\noindent\textbf{ECG (non-optimal)} replaces the LBG in GHOST with a complete edge-cost graph (ECG), assigning each abstract edge $(u,v)$ the Euclidean distance between the Chebyshev centers~\cite{boyd2004convex} of $\mathcal{X}_u$ and $\mathcal{X}_v$. ECG solves the original RTSP ILP over abstract edges~\cite{van1999solving} and unfolds tours using edge costs. It is non-optimal as these costs do not yield valid lower bounds.


\noindent\textbf{Greedy (non-optimal)} uses greedy search on $G$ with its LBG. Given an initial vertex, a depth-first search chooses the next unvisited vertex $v$ from the current vertex $u$ such that $p=(\cdot,u,v)\in\mathcal{P}$ has the smallest $lb_p$. 
After visiting all vertices, an abstract tour is unfolded via Alg.~\ref{alg:bfs}, and the search backtracks to explore other tours.
The best trajectory found is recorded throughout the greedy search. 

\noindent\textbf{MICP} solves GCS-TSP via an unified optimization~\cite{marcucci2024graphs}, assuming $G=(V,E)$ is complete.
For incomplete $G$, it is replaced with $|V|$ copies, denoted as $G^{i},i=1,2,...,|V|$. Each $G^{i}$ represents GCS $G$ at a different step $i$ and is connected to $G^{i+1}$ as described in~\citet{marcucci2024graphs}.

We generate GCS-TSP instances on a $5\times 5$ grid with $12$ random seeds (see an example in Fig.~\ref{fig:demo}-(a)), where the convex sets are centered at some intersections of the grid. The following three specific types of problems on Point-GCS, Linear-GCS, and Bézier-GCS, are used for evaluation.

\noindent\textbf{Point-GCS}: 
We generate complete GCSs of $N\in[5,25]$ vertices (convex sets).
Each convex set $\mathcal{X}_v\subseteq\mathbb{R}^2$ is a 2d point and each edge $e=(u,v)$ is unconstrained (i.e., $\mathcal{X}_e=\mathcal{X}_u\times \mathcal{X}_v$).
The trajectory cost $c(\tau)$ is defined as the sum of the Euclidean distances between consecutive points along $\tau$.

\noindent\textbf{Linear-GCS}:
We generate incomplete GCS of $M\in[10,50]$ edges.
Each convex set $\mathbf{x}_v\in\mathcal{X}_v\subseteq\mathbb{R}^{2+2}$ comprises of two points $\mathbf{a}_{v},\mathbf{b}_{v}\in\mathbb{R}^2$. Each edge $(u,v)$ indicates $\mathcal{X}_u\cap\mathcal{X}_v\neq\emptyset$ with edge constraint $\mathcal{X}_e:=\{(\mathbf{x}_u,\mathbf{x}_v)|\mathbf{b}_u=\mathbf{a}_v\}\subseteq \mathcal{X}_u\times\mathcal{X}_v$.
The trajectory cost $c(\tau)$ is defined as the sum of the Euclidean distance between $\mathbf{a}_v$ and $\mathbf{b}_v$ for each $v$ along $\tau$.

\noindent\textbf{Bézier-GCS}: 
We generate incomplete GCS of $M\in[10,50]$ edges. Each $\mathbf{x}_v\in\mathcal{X}_v\subseteq\mathbb{R}^{3\times 5}$ consists of $5$ control points for two $4$-order Bézier curves: one in 2D space and the other in 1D time that reconstruct a continuous trajectory piece bounded by $\mathcal{X}_v$.
An edge $(u,v)$ connects $\mathcal{X}_u$ and $\mathcal{X}_v$ if $\mathcal{X}_u\cap\mathcal{X}_v\neq\emptyset$, with $\mathcal{X}_e$ defined as trajectory continuity constraints (see~\citet{marcucci2023motion}).
The cost $c(\tau)$ is defined as the sum of the durations of each 1D time Bézier curve.

\subsection{Comparison, Ablation Study, and Demonstration}
\begin{figure}[t]
\centering
\includegraphics[width=\linewidth]{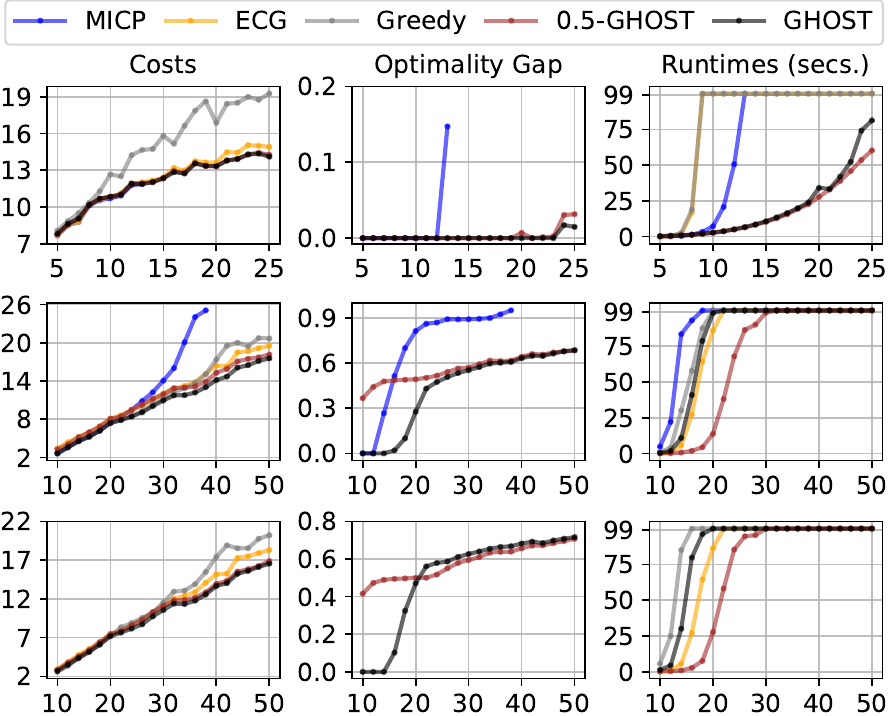}
\caption{Comparison and ablation study on Point-GCS, Linear-GCS, and Bézier-GCS (from top to bottom row), where their x-axes indicate the number of vertices, number of edges, and number of edges, respectively.}
\label{fig:exp}
\end{figure}

Fig.~\ref{fig:exp} summarizes the performance comparisons.

\noindent\textbf{Metrics:}
We compare all methods in terms of solution trajectory cost $c(\tau)$ and runtime.
Each method is given a $100$-second limit per random GCS-TSP instance, and metrics for each instance are averaged over $12$ random seeds.
We also report the optimality gap $\rho=(c(\tau^*)-lb^*)/c(\tau^*)$ for GHOST and MICP, where $lb^*$ is the greatest provable lower bound at termination.
For GHOST, $lb^*=\mathcal{L}(\hat\pi_i)$ is the cost of the next-best abstract tour when $c(\tau^*)$ is returned.
Note that $\mathcal{L}(\hat\pi_i)>c(\tau^*)$ is possible for an optimal $c(\tau^*)$, so negative $\rho$ values are capped at zero.
For MICP, $lb^*\leq c(\tau^*)$ is the current-best fractional solution cost in branch-and-bound.

\noindent\textbf{Result Analysis:} GHOST consistently outperforms ECG and Greedy in both trajectory cost and runtime.
Even without LBG, ECG is a strong baseline, as it uses the edge cost estimates to guide the GHOST search, highlighting the flexibility and generality of our GHOST framework. The comparison with Greedy, which replaces systematic search with a heuristic approach, further demonstrates the advantage of GHOST: Systematic exploration yields better solutions and more reliable performance.
A key strength of GHOST is the integration of LBG, which provides a provable lower-bound cost for the solution trajectory. This feature is particularly valuable in practice, as it allows the optimality gap to be monitored throughout the search, enabling users to gauge solution quality even if the search is terminated before optimality is achieved. In contrast, ECG and Greedy lack such guarantees (unless all tours are explicitly enumerated), leading to longer runtimes and no certificate of optimality---especially noticeable in the Point-GCS cases.
GHOST also shows clear scalability benefits over MICP, which fails to produce feasible solutions for Point-GCS with $N>13$, Linear-GCS with $M>19$, or any Bézier-GCS instances.
The bounded-suboptimal $0.5$-GHOST variant further highlights the flexibility of our approach, often obtaining solutions of the same quality as those of GHOST, but with improved efficiency---particularly when runtime is limited.

\noindent\textbf{Applications}:
We demonstrate the versatility of GHOST on Bézier-GCS-TSP instances for coverage, inspection, and task and motion planning.
Fig.~\ref{fig:teaser} and Fig.~\ref{fig:demo}-(c) show continuous trajectories covering all precomputed collision-free convex regions.
Fig.~\ref{fig:demo}-(b) shows inspection of a selected subset of the precomputed convex regions, achieved by modifying our RTSP ILP to require only coverage of the subset.
Finally, Fig.~\ref{fig:demo}-(d) demonstrates a 7-DoF manipulator (KUKA LBR iiwa) reaching $8$ independent convex sets, each corresponding to a distinct task pose.

\section{Conclusions and Future Work}
We presented GHOST, an optimal hierarchical framework for GCS-TSP that combines combinatorial tour search with convex trajectory optimization.
GHOST systematically explores tours on GCS while computing admissible lower bounds that guide best-first search at both the tour level and the path unfolding level for feasible trajectories realizing each tour, enabling effective pruning while maintaining optimality guarantees. 
Future work includes extending GHOST to general graph optimization problems in GCS, improving its scalability, and adapting it for multi-agent systems.

\section*{Acknowledgements}
This work was supported by the NSERC under grant number RGPIN2020-06540 and a CFI JELF award.

{\small
\bibliography{aaai2026}
}


\end{document}